\documentclass[runningheads]{llncs}
\usepackage{graphicx}
\usepackage{times}

\usepackage{soul}
\usepackage{url}
\usepackage[hidelinks]{hyperref}
\usepackage[utf8]{inputenc}
\usepackage{graphicx}
\usepackage{amsmath}
\usepackage{amsfonts}
\usepackage{booktabs}
\usepackage{float}
\begin{document}
\title{Graph Neural Networks for Propositional Model Counting}
%
%
\author{Gaia Saveri\inst{1,2} \and
Luca Bortolussi\inst{2}}
\authorrunning{G. Saveri et L. Bortolussi}

\institute{Department of Computer Science, University of Pisa, Italy \\ \email{gaia.saveri@phd.unipi.it} \and
Department of Mathematics and Geoscience, University of Trieste, Italy
}
\maketitle              
\begin{abstract}
Graph Neural Networks (GNNs) have been recently leveraged to solve several logical reasoning tasks. Nevertheless, counting problems such as propositional model counting (\#SAT) are still mostly approached with traditional solvers. Here we tackle this gap by presenting an architecture based on the GNN framework for belief propagation (BP) of \cite{kuck_belief_2020}, extended with self-attentive GNN and trained to approximately solve the \#SAT problem. We ran a thorough experimental investigation, showing that our model, trained on a small set of random Boolean formulae, is able to scale effectively to much larger problem sizes, with comparable or better performances of state of the art approximate solvers. Moreover, we show that it can be efficiently fine-tuned to provide good generalization results on different formulae distributions, such as those coming from SAT-encoded combinatorial problems.
\end{abstract}

\section{Introduction}\label{sec:intro}

Propositional model counting (\#SAT), the problem of computing the number of satisfying assignments of a given Boolean formula, is of relevant importance in computer science as it arises in many domains, such as Bayesian reasoning and combinatorial designs \cite{biere_handbook_2009}. Nevertheless, \#SAT is \#P-complete, thus computationally at least as hard as NP-complete problems \cite{valiant_complexity_1979}. \footnote{being \#P-complete the complexity class of counting problems associated with NP-complete decision problems.} 

For this reason, state of the art exact \#SAT solver are not capable of handling industrial-size problems, hence a number of approximate solvers (i.e. methods providing an estimation of the number of solutions) have been developed, which are able to scale to larger problem sizes. 

On the other hand, in the last few years there has been an increasing interest in leveraging machine learning in general and deep learning in particular to solve logical and combinatorial reasoning tasks \cite{li_combinatorial_2018,lamb_graph_2020,bengio_machine_2021}. Graph Neural Networks (GNNs) \cite{scarselli_graph_2009} fit well in this scenario as they carry inductive biases that effectively encode combinatorial inputs, such as permutation invariance and sparsity awareness \cite{battaglia_relational_2018}. 
Although not always providing the same exactness guarantees of traditional solvers, the rationale of using learning-based approaches (among which GNNs) in these fields is that they can provide a fast approxmation of the computations performed by non-learned solvers, without the need of specifying distribution-specific hand-crafted heuristics. 

In order to be practically exploited in combinatorial tasks, deep learning models should satisfy some desiderata \cite{cappart_combinatorial_2021}, that we address in the experimental evaluation of this work. These properties are: scalability (i.e. the model should be able to perform well on graph instances which are from the same data generating distribution than the ones seen during training, but of larger size), generalization (i.e. the model should perform well on unseen instances), data-efficiency (i.e. the model should not require a large amount of labeled data for training, as it may be costly to obtain) and time-efficiency (i.e. the performance of the model should be close to that of optimal solvers, but faster).

In this work we investigate whether GNNs can be meaningfully applied to approximately solve the \#SAT problem, an objective that, to the best of our knowledge, is only tackled in \cite{kuck_belief_2020}. To this end, we extend the architecture presented in \cite{kuck_belief_2020}, that aims at generalizing the Belief Propagation algorithm (BP) \cite{pearl_probabilistic_1988} using Message Passing Neural Networks (MPNNs) \cite{gilmer_neural_2017}, by augmenting the model with a self-attention mechanism. 

The rationale behind keeping the algorithmic structure of BP as in \cite{kuck_belief_2020} is that the estimate of the partition function given by BP has been proven to be a good approximation of the number of solutions of a Boolean formula \cite{kroc_leveraging_2011}, and also because the dynamics of MPNNs reflects the way in which information is propagated throughout the graphical model by BP. The motivation for introducing an attention mechanism inside the architecture is instead that of enhancing generalization capabilities of the model, while  preserving the relational inductive biases needed in this context. 

We carry out a careful and extensive experimental investigation of our proposed model, addressing in particular the already mentioned scalability and generalization properties. We show that our architecture, trained on a small set of synthetically generated random Boolean formulae, is able to scale-up to larger problem sizes, outperforming state-of-the art approximate \#SAT solver. Moreover we describe a simple yet effective fine-tuning strategy, that allows the model to generalize across diverse data distributions, with only a few tens of labeled formulae required. 

\section{Background}\label{sec:backgroung}

\subsection{Belief Propagation and \#SAT}\label{subsec:bp}

Belief Propagation \cite{pearl_probabilistic_1988} is an approximate inference algorithm for computing marginals of a probability distribution, exploiting the factor graph arising from its factorization. If we consider a discrete probability distribution over variables $V=\{x_1, \ldots, x_n\}$ (below $x_{\mathcal{N}(f_j)}$ indicates the set of variables each factor $f_j$ depends on):

\begin{equation}
\begin{split}
P(x_1,\ldots, x_n)  = \frac{1}{Z} \prod_{j=1}^m f_j(x_{\mathcal{N}(f_j)}), \text{  } Z = \sum_{x\in V} \prod_{j=1}^m f_j(x_{\mathcal{N}(f_j)})
\end{split}
\label{eq:prob}
\end{equation}
then belief propagation computes an approximation of the factor marginals (also called beliefs) $\{b_j(x_{\mathcal{N}(f_j)})\}_{j=1}^m$ and of the variable marginals $\{b_i(x_i)\}_{i=1}^n$ by passing messages between neighboring nodes on the factor graph following the iterative scheme given by:

\begin{equation}
\begin{split}
m_{i\rightarrow j}^{(k+1)}(x_i) &= \prod_{c\in \mathcal{N}(x_i)\backslash j} m_{c\rightarrow i}^{(k)}(x_i) \\
m_{j\rightarrow i}^{(k+1)}(x_i) &= \sum_{x_1,\ldots, x_k\in \mathcal{N}(f_j)\backslash x_i} f_j(x_i, x_1,\ldots, x_k) \cdot \prod_{x_v\in \mathcal{N}(f_j)\backslash x_i)} m_{v\rightarrow i}^{(k)}(x_v)
\end{split}
\label{eq:bp}
\end{equation}
with initialization $m^{(0)}_{i\rightarrow j}(x_i) = 1$ for variable-to-factor messages and $m^{(0)}_{j\rightarrow i}(x_i) = f_j(x_i)$ for factor-to-variable messages. When convergence has been reached, or a predefined maximum number of iterations $T$ has been performed, approximate marginals are computed. Beliefs can be used to compute a variational approximation of the partition function of the factor graph $Z$ of Equation \ref{eq:prob} as (below $deg(x_i)$ denotes the number of edges incident to variable node $x_i$):

\begin{equation}
\begin{split}
U &= -\sum_{j=1}^m \sum_{x_{\mathcal{N}(f_j)}} b_j(x_{\mathcal{N}(f_j)})\ln f_j(x_{\mathcal{N}(f_j)}) \\
H &= -\sum_{j=1}^m \sum_{x_{\mathcal{N}(f_j)}} b_j(x_{\mathcal{N}(f_j)}) \ln b_j(x_{\mathcal{N}(f_j)})  + \sum_{i=1}^n (deg(x_i)-1)\sum_{x_i} b_i(x_i) \ln b_i(x_i) \\
F &= U - H
\end{split}
\label{eq:partition}
\end{equation}
where $U$ is called Bethe average energy, $H$ Bethe entropy and $F$ Bethe free energy. It holds that $F\approx -\ln Z$ \cite{bethe_statistical_1935}. 
It is worth noting that, in order to avoid underflow errors when implementing BP, message updates of Equation \ref{eq:bp} are usually performed in the log-space. Moreover, it has been shown that taking partial updates improves convergence of BP without changing its fixed points \cite{kroc_leveraging_2011}. In detail, given a \textit{damping} parameter $\alpha\in [0,1]$, the following weighted combination is taken at each message passing iteration:

\begin{equation}
\begin{split}
m_{i\rightarrow j}^{(k+1)}(x_i) &=  \alpha\cdot m_{i\rightarrow j}^{(k+1)}(x_i) + (1-\alpha)\cdot m_{i\rightarrow j}^{(k)} \\
m_{j\rightarrow i}^{(k+1)}(x_i) &=  \alpha\cdot m_{j\rightarrow i}^{(k+1)}(x_i) + (1-\alpha)\cdot m_{j\rightarrow i}^{(k)} \\
\end{split}
\label{eq:damping}
\end{equation} 

It holds that propositional formulae, that without loss of generality we assume to be in Conjunctive Normal Form (CNF), can be readily translated into a factor graph. Such bipartite graph contains a factor node for every clause and a variable node for every variable in the formula, and undirected edges connecting variable nodes to the factor nodes of the clauses they appear in. If we impose that a factor node takes value $1$ for variable configurations that satisfy the corresponding clause and $0$ otherwise, then the partition function of this factor graph (Z of Equation \ref{eq:prob}) counts, by construction, the number of models of the input formula, i.e. the solution to the \#SAT problem (implementing this also allows to distinguish between two formulae differing only for literals’ negation, which in principle have the same factor graph representation). 
This intuition enables the adoption of probabilistic reasoning methods as approximate solvers for \#SAT \cite{kroc_leveraging_2011}. On SAT instances, BP works by iteratively passing messages between variables and clauses until a fixed point is reached. From the fixed point, an approximation of the partition function $Z$ is computed, which serves as an approximation to the number of models of the input formula. 

\subsection{Graph Attention Networks}\label{subsec:gat}

Graph Neural Networks \cite{scarselli_graph_2009} are deep learning models that address graph-related task, being natively able to process graph-structured inputs. The key idea behind GNNs is to compute a continuous representation of the nodes of the input graph that strongly depends on the structure of the graph. In particular, the Message Passing Neural Network model (MPNNs) \cite{gilmer_neural_2017}, which provides a general framework for GNNs, uses a form of neural message passing, in which real-valued vector messages are exchanged between neighboring nodes, for a fixed number of iterations, in such a way that at each iteration each vertex aggregates information from its immediate neighbors. 

Graph Attention Networks (GATs) \cite{velickovic_graph_2018} are a type of GNN endowed with a self-attention mechanism, that allows the network to aggregate the information coming from different nodes of the input graph putting different focus (i.e. a different weight) on some entities, and fade out the rest. More in detail, given an input graph $G=(V, E)$ and a set of continuous embeddings for the nodes of the graph  $\textbf{h}=\{\textbf{h}_1, \ldots, \textbf{h}_{|V|}\}$, with $\textbf{h}_i\in \mathbb{R}^d$, the graph attentional layer computes the attention coefficients between node $i$ and node $j$ as:

\begin{equation}
\alpha_{ij} = \frac{\exp(\text{LeakyReLU}(\textbf{a}^T[\text{concat}(W\textbf{h}_i, W \textbf{h}_j)]))}{\sum _{k\in \mathcal{N}(i)} \exp(\text{LeakyReLU}(\textbf{a}^T[\text{concat}(W\textbf{h}_i , W \textbf{h}_k)]))}
\label{eq:GAT_coef}
\end{equation}
for all pairs of neighboring nodes, where $W\in \mathbb{R}^{d\times d}$ is a learnable matrix and $\textbf{a}\in \mathbb{R}^{2d}$ is the set of parameters of a single-layer feedforward neural network. To update node representation and obtain a new set of node features $\textbf{h}^{'} = \{\textbf{h}_1^{'}, \ldots, \textbf{h}_{|V|}^{'}\}$ the following is computed:

\begin{equation}
\textbf{h}_i^{'} = \sigma \bigg(\sum_{j\in \mathcal{N}(i)} \alpha_{ij} W \textbf{h}_j \bigg)
\label{eq:GAT_update}
\end{equation}
being $\sigma$ a non-linear differentiable function. In order to stabilize the learning process of self-attention, a multi-head attention (similar to that of \cite{vaswani_attention_2017}) is applied by replicating the operations of the layer (Equations \ref{eq:GAT_coef} and \ref{eq:GAT_update}) independently $K$ times and aggregating the results as:

\begin{equation}
\textbf{h}_i^{'} = \bigg \|_{k=1}^K \sigma \bigg( \sum_{j\in \mathcal{N}(i)} \alpha_{ij}^k W^k \textbf{h}_j \bigg) 
\label{eq:GAT_multihead}
\end{equation}
where $\|$ denotes a feature-wise aggregation function such as concatenation, sum or average. 

\section{Method}\label{sec:method}

The objective of this work is to tackle the \#SAT problem using a GNN model as an approximate solver.

\subsection{Belief Propagation Neural Networks}\label{subsec:bpnn}
The starting point of our investigation is (one of the variants of) the architecture proposed in \cite{kuck_belief_2020}, that we recall here briefly. This model, called Belief Propagation Neural Network (BPNN), generalizes BP by means of GNNs, taking as input the factor graph representing a CNF SAT formula and giving as output an estimation of the logarithm of the factor graph's partition function $Z$ of Equation \ref{eq:prob}.  Given an input factor graph $G=(V, E)$, where nodes are partitioned into factor nodes $\{f_j\}_{j=1}^m$ and variable nodes $\{x_i\}_{i=1}^n$, both factor-to-variable messages $\hat{m}_{j\rightarrow i}$ and variable-to-factor $\hat{m}_{i\rightarrow j}$ messages are initialized to $1$; then the following message passing phase is performed (in the log-space) for $T$ iterations:

\begin{equation}
\begin{split}
\hat{m}^{(k+1)}_{i\rightarrow j} (x_i) &= \sum_{c\in \mathcal{N}(x_i)\backslash j} \text{MLP}_1(\hat{m}^{(k)}_{c\rightarrow i}(x_i)) \\
\hat{m}^{(k+1)}_{j\rightarrow i} (x_i) &= \text{LSE}_{x_1,\ldots, x_k\in \mathcal{N}(f_j)\backslash x_i} \bigg( f_j(x_i, x_1,\ldots, x_k) + \\
& + \sum_{x_v\in \mathcal{N}(f_j)\backslash x_i)} \text{MLP}_2(\hat{m}_{v\rightarrow i}^{(k)}(x_v)) \bigg)
\end{split}
\label{eq:bpnn_update}
\end{equation}
where $\text{LSE}$ is a shorthand for the log-sum-exp function. That is, BPNN augments the standard BP message passing scheme of Equation \ref{eq:bp} (in the log-space) by transforming the messages using MLPs. 

After the message passing phase is completed, the following readout phase is executed, which outputs an estimation $\ln \hat{Z}$ of the natural logarithm of the partition function $Z$ of the input factor graph:

\begin{equation}
\begin{split}
\ln \hat{Z} & = \text{MLP}_3\bigg[ \text{concat}_{k=1}^T \bigg( \text{concat} \bigg( \sum_{j=1}^m b_j^{(k)}(x_{\mathcal{N}(f_j)})\ln f_j(x_{\mathcal{N}(f_j)}),  \\
 - & \sum_{j=1}^m b_j^{(k)}(x_{\mathcal{N}(f_j)}) \ln b_j^{(k)}(x_{\mathcal{N}(f_j)}), \\
& \sum_{i=1}^n (\text{deg}(x_i)-1) b_i^{(k)}(x_i) \ln b_i^{(k)} (x_i) \bigg) \bigg)\bigg]
\end{split}
\label{eq:bpnn_readout}
\end{equation}
where $b_i^{(k)}(x_i)$ and $b_j^{(k)}(x_{\mathcal{N}(f_j)})$ refer to an approximation of variable and factor beliefs at iteration $k$, respectively, computed as in standard belief propagation (we defer to the supplementary material, Section \ref{app:back} for further details).
Hence this final layer takes as input a concatenation of a term representing the Bethe average energy ($U$ of Equation \ref{eq:partition}) and two terms representing the components of the Bethe entropy ($H$ of Equation \ref{eq:partition}) across all iterations, summed across factors within each iteration. 
As already mentioned in Section \ref{subsec:bp}, damping is a standard technique for improving convergence of BP. The BPNN architecture allows for possibly applying a learned operator $\Delta: \mathbb{R}^{\sum_{i=1}^n 2\cdot deg(x_i)} \rightarrow \mathbb{R}^{\sum_{i=1}^n 2\cdot deg(x_i)}$ to the difference between iterations $k+1$ and $k$ of every factor-to-variable message, and jointly modify it, in place of the scalar multiplier $\alpha$ of Equation \ref{eq:damping}.

\subsection{Neural Belief Propagation with Attention}\label{subsec:bpgat}

Attention mechanisms, as seen in Section \ref{subsec:gat}, are a technique that allows a deep learning model to combine input representations additively, while also enforcing permutation invariance. GNNs endowed with an attention mechanism are a promising research direction in the neuro-symbolic computing scenario \cite{bengio_machine_2021,kool_attention_2019,lamb_graph_2020}, as they might enhance structured reasoning and efficient learning.

This is the reason for modifying the BPNN architecture by augmenting it with a GAT-style attention mechanism (i.e. analogous to that of Equations \ref{eq:GAT_coef}, \ref{eq:GAT_update}, \ref{eq:GAT_multihead}). 
In what follows we will refer to our architecture as BPGAT, to underline the fact that it puts together the algorithmic structure of BP and the computational formalism of GATs.

BPGAT works by taking as input a bipartite factor graph $G=(V, E)$, with factor nodes $\{f_j\}_{j=1}^m$ and variable nodes $\{x_i\}_{i=1}^n$, representing a CNF SAT formula and outputs an approximation $\ln \hat{Z}$ of the logarithm of the exact number of solutions of the input formula. As for BPNN, messages between nodes are partitioned into factor-to-variable messages and variable-to-factor messages, and computations are performed in the log-space.

Considering variable-to-factor messages, at each iteration $k+1$ every variable node $x_i$ contains the aggregated messages $\hat{m}_{i\rightarrow j}^{(k)}$ received from its neighborhoods at the previous iteration $k$, and needs to receive messages from its adjacent clause nodes, and aggregate them. Such aggregation is done using a GAT-style multi-head attention mechanism: as for the GAT attentional layer, the model computes attention coefficients $\alpha_{ij}$ for every factor node $f_j$ in the neighborhood of the variable node $x_i$ by projecting each message $\hat{m}^{(k)}_{j\rightarrow i}\in \mathbb{R}^2$ and variable node hidden representation representation $\hat{m}_{i\rightarrow j}^{(k)}\in \mathbb{R}^2$ via a weight matrix $W\in \mathbb{R}^{2\times 2}$, then passing the concatenation of the projected messages through a single-layer feedforward neural network $\textbf{a}\in \mathbb{R}^4$, then feeding the results to a LeakyReLU and finally applying softmax to normalize them across the neighborhood $\mathcal{N}(x_i)$ of $x_i$. More formally, attention coefficients are computed as:

\begin{equation}
    \displaystyle
    \alpha_{ij}^{(k+1)} = \frac{\exp(\text{LeakyReLU}(\textbf{a}^T[\text{concat}(W\hat{m}_{i\rightarrow j}^{(k)}, W \hat{m}^{(k)}_{j\rightarrow i} )]))}{\sum _{k\in \mathcal{N}(x_i)} \exp(\text{LeakyReLU}(\textbf{a}^T[\text{concat}(W\hat{m}_{i\rightarrow j}^{(k)}, W \hat{m}^{(k)}_{k\rightarrow i})]))}
\label{eq:bp_att_coef}
\end{equation}

The new variable-to-factor messages are then computed as a weighted sum of the incoming factor-to-variable messages, using attention coefficients of Equation \ref{eq:bp_att_coef}, as:
\begin{equation}
\hat{m}_{i\rightarrow j}^{(k+1)}(x_i) = \sum_{c\in \mathcal{N}(x_i)\backslash j} \alpha_{ij}^{(k+1)} W \hat{m}_{c\rightarrow i}^{(k)}(x_i)
\label{eq:var2fact}
\end{equation}

In order to stabilize the learning process, the multi-head attention mechanism of Equation \ref{eq:GAT_multihead} is used, by replicating $K$ times the computations of Equations \ref{eq:bp_att_coef} and \ref{eq:var2fact} and aggregating the results. To perform the aggregation of the outputs of each of the attention heads, concatenation has been used for the internal layers, average for the final one.

Analogous computations are performed to compute factor-to-variable messages:
\begin{equation}
\begin{split}
\alpha_{ji}^{(k+1)} & = \frac{\exp(\text{LeakyReLU}(\textbf{a}^T[\text{concat}(W\hat{m}_{j\rightarrow i}^{(k)}, W \hat{m}^{(k)}_{i\rightarrow j} )]))}{\sum _{x_v\in \mathcal{N}(f_j)} \exp(\text{LeakyReLU}(\textbf{a}^T[\text{concat}(W\hat{m}_{j\rightarrow i}^{(k)}, W \hat{m}^{(k)}_{i\rightarrow v})]))} \\
\hat{m}_{j\rightarrow i}^{(k+1)}(x_i) & = \text{LSE}_{x_1,\ldots, x_k\in \mathcal{N}(f_j)\backslash x_i} \bigg( f_j(x_i, x_1,\ldots, x_k) + \\ 
& + \sum_{x_v\in \mathcal{N}(f_j)\backslash x_i} \alpha_{ji}^{(k+1)} W \hat{m}_{v\rightarrow i}^{(k)}(x_i) \bigg)
\end{split}
\label{eq:fact2var}
\end{equation}

Also for this type of messages, a multi-head attention mechanism is deployed, where the output of each head is concatenated in internal layers and averaged in the final layer.

Once this message passing phase has run for $T$ iterations, the readout phase is executed. This consists in applying as final layer the one described in Equation \ref{eq:bpnn_readout}, in order to obtain $\ln \hat{Z}$, i.e. an approximation of the natural logarithm of the number of models of the CNF SAT formula encoded by the input factor graph $G$. 

\section{Experimental Evaluation}\label{sec:experiments}

\subsection{Experimental Setting}\label{subsec:setting}

We implemented the BPGAT architecture in Python, leveraging the PyTorch framework \cite{paszke_pytorch_2019}. The model is trained to minimize the Mean Squared Error (MSE) between the natural logarithm $\ln Z$ of the true number of models of the input formula, and the output of the model $\ln \hat{Z}$. 

\subsubsection{BPGAT Training Protocol}\label{subsec:bpgat_train}

We trained the model for $1000$ epochs using the Adam optimizer \cite{kingma_adam_2015} with an initial learning rate of $10^{-4}$, halving it every $200$ epochs using a learning rate scheduler. 

Given an input formula $\phi$ with $n$ variables $\{x_i\}_{i=1}^n$ and $m$ clauses $\{f_j\}_{j=1}^m$ and its factor graph representation $G=(V,E)$, it is preprocessed before being fed to the network in such a way that $\forall j\in \{1, \ldots, m\}, \forall \{x_1, \ldots, x_k\}\in \mathcal{N}(f_j), f_j(x_1, \ldots, x_k)=1$ for the assignment of $\{x_1, \ldots, x_k\}$ that makes clause $f_j$ evaluate to true, $0$ otherwise, to ensure that $Z$ of Equation \ref{eq:prob} actually represents the count of models satisfying $\phi$.

In order to compute attention coefficients of Equations \ref{eq:var2fact} and \ref{eq:fact2var}, we used a $3$-layer GAT network, having respectively $4, 4, 6$ attention heads. Moreover, we used a $3$-layer MLP $\Delta$, with ReLU activation between hidden layers, to transform factor-to-variable messages, in place of a fixed-scalar damping parameter.  For variable-to-factor messages a damping parameter $\alpha=0.5$ has been used. The MLP of the final layer (Equation \ref{eq:bpnn_readout}) is a $3$-layer feedforward network, with ReLU non-linerity between hidden layers. The number of iterations $T$ of the message passing scheme has been set to $5$. This architectural choices have been made after performing a set of preliminary and ablation studies, whose results can be found in supplementary material, Section \ref{subapp:par_tuning}.

\subsubsection{BPGAT Training Data}\label{subsec:train_data}
The training dataset $\mathcal{D}=\{(\phi_i, \ln Z_i)\}$ consists of a set of $1000$ pairs of CNF SAT formulae $\phi_i$ and the logarithm $\ln Z_i$ of their true model count. Such formulae are drawn from a distribution of random formulae built in the following way:
\begin{enumerate}
\item Given input parameters $(nv_{min}, nv_{max})$, the number of variables of the current formula $\phi$ is chosen as $n_{var} \sim \mathcal{U}([nv_{min}, nv_{max}])$.
\item Given input parameters $(nc_{min}, nc_{max})$, the number of clauses of $\phi$ is chosen as $n_{cl}\sim \mathcal{U}([nc_{min}, nc_{max}])$.
\item For each clause $c_j$, with $j\in \{1, \ldots, n_{cl}\}$, a number $k\sim 2+\text{Bernoulli}(0.7)+\text{Geometric}(0.4)$ of variables are chosen uniformly at random from the input $n_{var}$ variables; this leads to clause having $5$ variables in average. Once the variables for the current clause $c_j$ are chosen, each of them is negated with probability $0.5$; before adding $c_j$ to the formula $\phi$, it is checked that $c_j$ is different from all the other clauses already in the formula.
\item As soon as all the $n_{cl}$ clauses are generated, the resulting formula $\phi$ is fed to Minisat \cite{een_extensible_2003} and checked for satisfiability: it is added to the dataset $\mathcal{D}$ if and only if it is SAT.
\item All formulae $\phi_i$ in the dataset $\mathcal{D}$ are fed to sharpSAT \cite{thurley_sharpsat_2006}, an exact \#SAT solver to generate the true count $Z_i$, and add its logarithm $\ln Z_i$ to the dataset. 
\end{enumerate}
The dataset used for training BPGAT has been generated using $(nv_{min}, nv_{max})=(10, 30)$ and  $(nc_{min}, nc_{max})=(20, 50)$; this produced formulae having an average of $19.87$ variables and $34.87$ clauses. 

\subsubsection{Fine-tuning Protocol}\label{subsec:fine_tune}
In order to allow the model to extrapolate to data distributions different from the one seen during training, without requiring a large amount of labeled data (as ground-truth labels might be costly to obtain), we use a pre-trained BPGAT as weight initializer for fine-tuning towards new formulae distributions.  

In particular, we initialize the weights of our model using those of BPGAT trained for $500$ epochs with the protocol described earlier in this Section and data drawn from the previously proposed random distribution, and we train it for $250$ epochs with a learning rate of $10^{-6}$, with only $250$ labeled examples. 

\subsection{Results}\label{subsec:results}
The objective of our experiments is twofold: evaluating both BPGAT scalability and generalization capabilities. In order to asses the performance of the model, both Root Mean Squared Error (RMSE) and Mean Relative Error (MRE) metrics are reported, evaluated between the logarithm of the ground truth number of models of the input formulae $\ln Z$ and the output of the model $\ln \hat{Z}$. 

As a baseline, we used ApproxMC \cite{CMV16,SM19}, the state-of-the-art approximate \#SAT solver, which is a randomized hashing algorithm that provides Probably Approximately Correct (PAC) guarantees. 

It would have been meaningful and interesting to compare BPGAT against other guarantee-less counters, such as ApproxCount \cite{approxcount} or SampleApprox \cite{sampleapprox}, but unfortunately we couldn't access any open-source implementation of them. 

\subsubsection{Scalability} 
In order to assess the ability of our model to scale to larger problem sizes than the one seen during training, we generated several datasets following the procedure detailed in Section \ref{subsec:bpgat_train}.

Table \ref{tab:scala_stats} shows the statistics of the datasets used in this testing phase, each containing $300$ labeled instances. All datasets are much larger than the one seen during training, and in particular `Test 4` contains formulae having a number of variables which is more than ten times more the one in the training set, and a number of clauses which is almost ten times more than the ones in the training set.

Table \ref{tab:scala_res} shows the results obtained, in terms of RMSE and MRE, by BPGAT and ApproxMC. 

\begin{figure}
    \centering
    \includegraphics[width=0.8\textwidth, keepaspectratio]{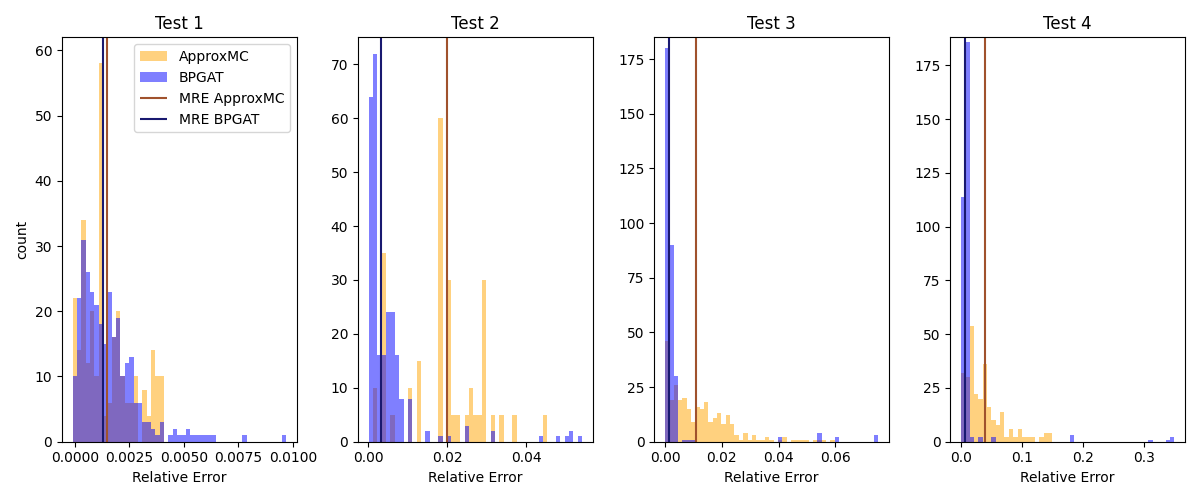}
    \caption{Distribution of the relative errors of BPGAT and ApproxMC on datasets used to test scalability.}
    \label{fig:mre_hist}
\end{figure}

It is worth noting that for all the datasets tested, BPGAT outperforms ApproxMC in terms of MRE (although not in terms of RMSE). Such higher RMSE is a consequence of few outliers with a large prediction error for BPGAT (as shown in Figure \ref{fig:mre_hist}), while most of its predictions are close to the ground truth labels, as certified  by the consistently lower MRE.

\begin{table}
\caption{Average number of variables, average number of clauses, average number of solutions and average time employed (in seconds) by the exact solver sharpSAT of the datasets used to test scalability.}
\label{tab:scala_stats}
\centering
\begin{tabular}{lrrrr}
\toprule
Dataset  & Avg\#var  & Avg\#cl & Avg\#sol & Avg t (s) \\
\midrule
Test 1    & 61.8  & 76.89 & 1.76e+19 & 26.71      \\
Test 2    & 60.43 & 143.61 & 6.23e+14 & 211.45     \\
Test 3    & 124.07 & 75.26 & 1.14e+21 & 28.5   \\ 
Test 4    & 377.59 & 275.11 & 7.18e+145 & 286.95    \\
\bottomrule
\end{tabular}
\end{table}

\begin{table}
\caption{RMSE/MRE performance of BPGAT and ApproxMC on datasets used to test scalability.}
\label{tab:scala_res}
\centering
\begin{tabular}{lrr}
\toprule
Dataset  & BPGAT  & ApproxMC \\
\midrule
Test 1    & 0.1276/\textbf{0.001366}  & \textbf{0.002025}/0.001576      \\
Test 2    & 0.3100/\textbf{0.003201}  & \textbf{0.2262}/0.02003      \\
Test 3    & 0.1748/\textbf{0.001471}  & \textbf{0.1035}/0.01134      \\
Test 4    & 1.2061/\textbf{0.007433}  & \textbf{0.4275}/0.04038     \\
\bottomrule
\end{tabular}
\end{table}

\subsubsection{Out-of-distribution generalization}
The second set of tests we performed aims at evaluating the generalization capabilities of our model. The problem classes we perform experiments with are both SAT-encoded combinatorial problems ($k$-dominating set, graph $k$-coloring, $k-$clique detection) and network QMR (Quick Medical Reference) problems taken from the test suite of \cite{SM19}.

In order to obtain SAT CNF formulae encoding the $k$-dominating set problem, the graph-$k$-coloring problem and the $k$-clique detection problem, we use the CNFgen tool \cite{lauria_cnfgen_2017}. It allows to sample graphs uniformly at random from the Erdos-Renyi random graph distribution $G(N, p)$  (specifying the two parameters $N>0, p\in[0, 1]$) and to encode in CNF the required problem. After obtaining the CNF formualae, Minisat \cite{een_extensible_2003} is used to filter SAT formulae and sharpSAT \cite{thurley_sharpsat_2006} to obtain the number of models for each formula. Table \ref{tab:gener_stats} shows the statistics of the datasets used in this testing phase.

To asses the effectiveness of our fine-tuning protocol we made the following experiments (whose results are summarized in Table \ref{tab:gener_exp}):
\begin{itemize}
    \item We fine-tuned the model following the protocol described in Section \ref{subsec:bpgat_train}. This is denoted as FT\_BPGAT in Table \ref{tab:gener_exp};
    \item We trained, for every distribution, the model from scratch for $500$ epochs, with $250$ labeled examples. This is denoted as TS\_BPGAT in Table \ref{tab:gener_exp};
    \item We tested BPGAT trained on random Boolean formulae (with the training protocol and the data generating procedures detailed in Section \ref{subsec:bpgat_train}). This is denoted as BPGAT in Table \ref{tab:gener_exp}.
\end{itemize}

As expected, the worst performance is achieved by BPGAT, as it has never been trained on formulae coming from these distributions. Interestingly, fine-tuning a model pre-trained on small random Boolean formulae (FT\_BPGAT) gives better results than the same model trained on the specific dataset (TS\_BPGAT). This is relevant from the perspective of data-efficiency because small random formulae are fast to generate (differently from other distributions) and the fine-tuning phase requires only a few tens of distribution-specific labeled samples. These results are also relevant from the time-efficiency perspective, as only retraining the model for $250$ epochs is needed, for each unseen data distribution.  

Results of the comparison between our fine-tuned model (FT\_BPGAT) and ApproxMC are shown in Table \ref{tab:gener_comp}. It is worth observing that the performance of our architecture is comparable and in some cases outperforming that of ApproxMC (especially in terms of MRE).

\begin{table}
\caption{Statistics of the datasets used to test generalization. Parameters under `Domset` (which stands for dominating set), `Color` (which stands for graph coloring) and `Clique` (which stands for clique detection) are the pair $(N, p)$ used to generate formulae in the dataset. For all of them $k=3$ has been used.}
\label{tab:gener_stats}
\centering
\begin{tabular}{lrrrr}
\toprule
Dataset  & Avg \# var  & Avg \# cl & Avg \# sol & Avg t (s) \\
\midrule
Network    & 113.3  & 294.7 & 1.19e+20 & 123.9      \\
Domset ($15, 0.6$)  & 38.43 & 510.15 & 203.3 & 3.76e-2     \\ 
Color ($10, 0.6$)   & 65.63 & 294.54 & 572.9 & 3.19e-2  \\
Clique ($15, 0.5$)  & 46.37 & 1145.73 & 102.2 & 3.86e-2   \\
\bottomrule
\end{tabular}
\end{table}

\begin{table}
\caption{RMSE/MRE performance of BPGAT when fine-tuned for the specific data distribution (FT\_BPGAT), when trained on the specific data distribution (TS\_BPGAT) and when trained on random formulae (BPGAT).}
\label{tab:gener_exp}
\centering
\begin{tabular}{lrrr}
\toprule
Dataset  & FT\_BPGAT  & TS\_BPGAT & BPGAT \\
\midrule
Network    & \textbf{0.2580}/\textbf{0.005271}  & 1.9334/0.04887  &  14.2839/0.3608 \\
Domset    & \textbf{0.5508}/\textbf{0.04252}  & 1.7808/0.7125  &  11.9190/9.5070 \\
Color    & \textbf{1.2110}/\textbf{0.1774}  & 1.3430/0.2046  &  26.1898/5.9593   \\
Clique    & \textbf{0.007834}/\textbf{0.002475}   & 0.01773/0.007333  & 1.9625/0.8983  \\
\bottomrule
\end{tabular}
\end{table}

\begin{table}
\caption{RMSE/MRE comparison of BPGAT fine-tuned for the specific data distributions (FT\_BPGAT) and ApproxMC.}
\label{tab:gener_comp}
\centering
\begin{tabular}{lrr}
\toprule
Dataset  & FT\_BPGAT & ApproxMC \\
\midrule
Network    & 0.2580/\textbf{0.005271}  & \textbf{0.07619}/0.05403 \\
Domset    & 0.5508/0.04252  & \textbf{0.08155}/\textbf{0.02856} \\
Color    & 1.2110/0.1774  & \textbf{0.09426}/\textbf{0.04241}  \\
Clique    & \textbf{0.007834}/\textbf{0.002475}   & 0.01113/0.04795  \\
\bottomrule
\end{tabular}
\end{table}

\subsubsection{Data and Time-efficiency} 

The BPGAT architecture can be claimed data-efficient, as it requires only $1000$ CNF small random formulae for (pre)training. Generating such training set requires $\sim 5s$ with the procedure described in Section \ref{subsec:bpgat_train}.

\begin{figure}
    \centering
    \includegraphics[width=0.6\textwidth, keepaspectratio]{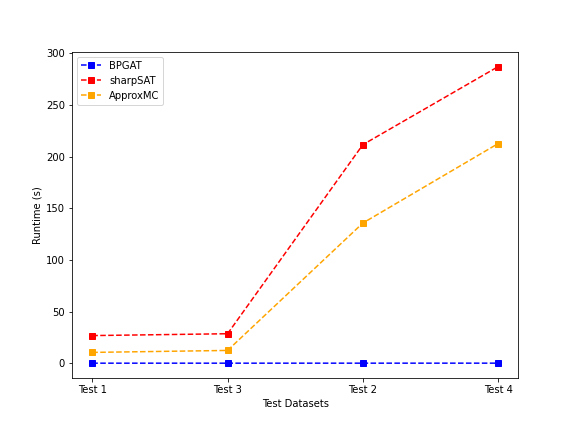}
    \caption{Runtime comparison (in seconds) of the mean time for counting the number of satisfying assignments of a formula for the datasets used in the scalability experiments.}
    \label{fig:runtime}
\end{figure}

For what concerns time efficiency, the BPGAT architecture is able to process (independently on the size of the formulae), all test instances described in this work, in a maximum of $3s$, without leveraging GPU acceleration. This is much less than the time needed by the exact solver sharpSAT and by the approximate solver ApproxMC, as shown in Figure \ref{fig:runtime}.

\section{Related Work}

In \cite{ACL-AAAI20} a graph neural network model is proposed to solve the weighted disjunctive normal form counting problem (weighted \#DNF). Differently from CNF \#SAT, there is  an approximate algorithm that admits a fully polynomial randomized approximation scheme (FPRAS), which allows the  generation of  hundreds of thousands of training data points, as opposed to our method which is forced to learn in a limited data regime. 

A quite significant body of work has been recently developed to learn how to solve $NP$-complete combinatorial optimization problems leveraging graph neural networks \cite{bengio_machine_2021}. Among them, methods tackling the Boolean satisfiability problem (SAT) are of particular interest for this work. They are divided into approaches leveraging GNNs as end-to-end solvers \cite{li_combinatorial_2018,selsam_learning_2019,amizadeh_learning_2019} (i.e. trained to output the solution directly from the input instance), and approaches in which the network is used as a computational tool to learn data-driven heuristics, in the loop of modern branch and bound solvers \cite{yolcu_learning_2019,kurin_can_2020}.

As already mentioned, our architecture builds upon the model of \cite{kuck_belief_2020}. Besides the architectural differences that have been underlined in Section \ref{sec:method}, the main distinction between the two models resides in the training protocol. Indeed, in \cite{kuck_belief_2020} training data consist in a sample of formulae drawn from the benchmarks used to test the architecture. Our model is instead trained on a set of random formulae, which are very fast to obtain, and then eventually fine-tuned towards the specific test distribution. 
The architectural improvements guarantee a sensibly better performance in terms of scalability and generalizability than the BPNN model, as shown in Section \ref{app:comp_bpnn} of the supplementary material. We believe that the attentional layer allows the network to focus on regions of the input formulae which are more significant for the \#SAT problem.

\section{Conclusions}
We presented BPGAT, an extension of the BPNN architecture presented in \cite{kuck_belief_2020}, which combines the algorithmic structure of belief propagation and the learning paradigm of graph attention networks. We conducted several experiments to investigate the scalability and generalization abilities of our network, showing that it is able to achieve a performance comparable to (and in some settings higher than) state-of-the-art approximate \#SAT solvers, albeit the lack of any theoretical guarantees on the quality of the solution. Finally, we highlighted the efficiency of our model, both in terms of required training data and in terms of running time.  

As future research directions, we will analyze the viability of extending our model to tackle weighted conjunctive normal form model counting (weighted \#CNF) problems. In this scenario, a straightforward application of our model would be that of approximate probabilistic inference on Bayesian Networks, which in many cases (e.g. when solved using variational inference) comes without any statistical guarantees. It would be also interesting to generalize this approach to other counting problems, such as counting the number of independent sets of a given size in a graph, as part of a broader research program aiming at exploring the application of graph attention networks to logical reasoning tasks.

\bibliographystyle{splncs04}
\bibliography{biblio_complete}

\begin{thebibliography}{10}
\providecommand{\url}[1]{\texttt{#1}}
\providecommand{\urlprefix}{URL }
\providecommand{\doi}[1]{https://doi.org/#1}

\bibitem{lauria_cnfgen_2017}
Cnfgen: {A} generator of crafted benchmarks. In: Gaspers, S., Walsh, T. (eds.)
  Theory and Applications of Satisfiability Testing - {SAT} 2017 - 20th
  International Conference, Melbourne, VIC, Australia, August 28 - September 1,
  2017, Proceedings. Lecture Notes in Computer Science, vol. 10491, pp.
  464--473. Springer (2017). \doi{10.1007/978-3-319-66263-3\_30},
  \url{https://doi.org/10.1007/978-3-319-66263-3\_30}

\bibitem{ACL-AAAI20}
Abboud, R., Ceylan, {\.I}.{\.I}., Lukasiewicz, T.: Learning to reason:
  Leveraging neural networks for approximate {DNF} counting. In: The
  Thirty-Fourth {AAAI} Conference on Artificial Intelligence, {AAAI} 2020, The
  Thirty-Second Innovative Applications of Artificial Intelligence Conference,
  {IAAI} 2020, The Tenth {AAAI} Symposium on Educational Advances in Artificial
  Intelligence, {EAAI} 2020, New York, NY, USA, February 7-12, 2020. pp.
  3097--3104. {AAAI} Press (2020),
  \url{https://ojs.aaai.org/index.php/AAAI/article/view/5705}

\bibitem{amizadeh_learning_2019}
Amizadeh, S., Matusevych, S., Weimer, M.: Learning to solve circuit-sat: An
  unsupervised differentiable approach. In: 7th International Conference on
  Learning Representations, {ICLR} 2019, New Orleans, LA, USA, May 6-9, 2019.
  OpenReview.net (2019), \url{https://openreview.net/forum?id=BJxgz2R9t7}

\bibitem{battaglia_relational_2018}
Battaglia, P.W., Hamrick, J.B., Bapst, V., Sanchez{-}Gonzalez, A., Zambaldi,
  V.F., Malinowski, M., Tacchetti, A., Raposo, D., Santoro, A., Faulkner, R.,
  G{\"{u}}l{\c{c}}ehre, {\c{C}}., Song, H.F., Ballard, A.J., Gilmer, J., Dahl,
  G.E., Vaswani, A., Allen, K.R., Nash, C., Langston, V., Dyer, C., Heess, N.,
  Wierstra, D., Kohli, P., Botvinick, M.M., Vinyals, O., Li, Y., Pascanu, R.:
  Relational inductive biases, deep learning, and graph networks. CoRR
  \textbf{abs/1806.01261} (2018), \url{http://arxiv.org/abs/1806.01261}

\bibitem{bengio_machine_2021}
Bengio, Y., Lodi, A., Prouvost, A.: Machine learning for combinatorial
  optimization: {A} methodological tour d'horizon. Eur. J. Oper. Res.
  \textbf{290}(2),  405--421 (2021). \doi{10.1016/j.ejor.2020.07.063},
  \url{https://doi.org/10.1016/j.ejor.2020.07.063}

\bibitem{bethe_statistical_1935}
Bethe, H.A.: Statistical {Theory} of {Superlattices}. Proceedings of the Royal
  Society of London. Series A, Mathematical and Physical Sciences
  \textbf{150}(871),  552--575 (1935)

\bibitem{biere_handbook_2009}
Biere, A., Heule, M., Maaren, H.v., Walsh, T. (eds.): Handbook of
  {Satisfiability}, Frontiers in {Artificial} {Intelligence} and
  {Applications}, vol.~185. IOS Press (2009)

\bibitem{cappart_combinatorial_2021}
Cappart, Q., Chetelat, D., Khalil, E.B., Lodi, A., Morris, C., Velickovic, P.:
  Combinatorial optimization and reasoning with graph neural networks. CoRR
  \textbf{abs/2102.09544} (2021), \url{https://arxiv.org/abs/2102.09544}

\bibitem{CMV16}
Chakraborty, S., Meel, K.S., Vardi, M.Y.: Algorithmic improvements in
  approximate counting for probabilistic inference: From linear to logarithmic
  {SAT} calls. In: Kambhampati, S. (ed.) Proceedings of the Twenty-Fifth
  International Joint Conference on Artificial Intelligence, {IJCAI} 2016, New
  York, NY, USA, 9-15 July 2016. pp. 3569--3576. {IJCAI/AAAI} Press (2016),
  \url{http://www.ijcai.org/Abstract/16/503}

\bibitem{een_extensible_2003}
E{\'{e}}n, N., S{\"{o}}rensson, N.: An extensible sat-solver. In: Giunchiglia,
  E., Tacchella, A. (eds.) Theory and Applications of Satisfiability Testing,
  6th International Conference, {SAT} 2003. Santa Margherita Ligure, Italy, May
  5-8, 2003 Selected Revised Papers. Lecture Notes in Computer Science,
  vol.~2919, pp. 502--518. Springer (2003).
  \doi{10.1007/978-3-540-24605-3\_37},
  \url{https://doi.org/10.1007/978-3-540-24605-3\_37}

\bibitem{gilmer_neural_2017}
Gilmer, J., Schoenholz, S.S., Riley, P.F., Vinyals, O., Dahl, G.E.: Neural
  message passing for quantum chemistry. In: Precup, D., Teh, Y.W. (eds.)
  Proceedings of the 34th International Conference on Machine Learning, {ICML}
  2017, Sydney, NSW, Australia, 6-11 August 2017. Proceedings of Machine
  Learning Research, vol.~70, pp. 1263--1272. {PMLR} (2017),
  \url{http://proceedings.mlr.press/v70/gilmer17a.html}

\bibitem{kingma_adam_2015}
Kingma, D.P., Ba, J.: Adam: {A} method for stochastic optimization. In: Bengio,
  Y., LeCun, Y. (eds.) 3rd International Conference on Learning
  Representations, {ICLR} 2015, San Diego, CA, USA, May 7-9, 2015, Conference
  Track Proceedings (2015), \url{http://arxiv.org/abs/1412.6980}

\bibitem{kool_attention_2019}
Kool, W., van Hoof, H., Welling, M.: Attention, learn to solve routing
  problems! In: 7th International Conference on Learning Representations,
  {ICLR} 2019, New Orleans, LA, USA, May 6-9, 2019. OpenReview.net (2019),
  \url{https://openreview.net/forum?id=ByxBFsRqYm}

\bibitem{kroc_leveraging_2011}
Kroc, L., Sabharwal, A., Selman, B.: Leveraging belief propagation, backtrack
  search, and statistics for model counting. Ann. Oper. Res.  \textbf{184}(1),
  209--231 (2011)

\bibitem{kuck_belief_2020}
Kuck, J., Chakraborty, S., Tang, H., Luo, R., Song, J., Sabharwal, A., Ermon,
  S.: Belief propagation neural networks. In: Larochelle, H., Ranzato, M.,
  Hadsell, R., Balcan, M., Lin, H. (eds.) Advances in Neural Information
  Processing Systems 33: Annual Conference on Neural Information Processing
  Systems 2020, NeurIPS 2020, December 6-12, 2020, virtual (2020)

\bibitem{kurin_can_2020}
Kurin, V., Godil, S., Whiteson, S., Catanzaro, B.: Can q-learning with graph
  networks learn a generalizable branching heuristic for a {SAT} solver? In:
  Larochelle, H., Ranzato, M., Hadsell, R., Balcan, M., Lin, H. (eds.) Advances
  in Neural Information Processing Systems 33: Annual Conference on Neural
  Information Processing Systems 2020, NeurIPS 2020, December 6-12, 2020,
  virtual (2020),
  \url{https://proceedings.neurips.cc/paper/2020/hash/6d70cb65d15211726dcce4c0e971e21c-Abstract.html}

\bibitem{lamb_graph_2020}
Lamb, L.C., d'Avila Garcez, A.S., Gori, M., Prates, M.O.R., Avelar, P.H.C.,
  Vardi, M.Y.: Graph neural networks meet neural-symbolic computing: {A} survey
  and perspective. In: Bessiere, C. (ed.) Proceedings of the Twenty-Ninth
  International Joint Conference on Artificial Intelligence, {IJCAI} 2020. pp.
  4877--4884. ijcai.org (2020). \doi{10.24963/ijcai.2020/679},
  \url{https://doi.org/10.24963/ijcai.2020/679}

\bibitem{li_combinatorial_2018}
Li, Z., Chen, Q., Koltun, V.: Combinatorial optimization with graph
  convolutional networks and guided tree search. In: Bengio, S., Wallach, H.M.,
  Larochelle, H., Grauman, K., Cesa{-}Bianchi, N., Garnett, R. (eds.) Advances
  in Neural Information Processing Systems 31: Annual Conference on Neural
  Information Processing Systems 2018, NeurIPS 2018, December 3-8, 2018,
  Montr{\'{e}}al, Canada. pp. 537--546 (2018),
  \url{https://proceedings.neurips.cc/paper/2018/hash/8d3bba7425e7c98c50f52ca1b52d3735-Abstract.html}

\bibitem{paszke_pytorch_2019}
Paszke, A., Gross, S., Massa, F., Lerer, A., Bradbury, J., Chanan, G., Killeen,
  T., Lin, Z., Gimelshein, N., Antiga, L., Desmaison, A., K{\"{o}}pf, A., Yang,
  E.Z., DeVito, Z., Raison, M., Tejani, A., Chilamkurthy, S., Steiner, B.,
  Fang, L., Bai, J., Chintala, S.: Pytorch: An imperative style,
  high-performance deep learning library. In: Wallach, H.M., Larochelle, H.,
  Beygelzimer, A., d'Alch{\'{e}}{-}Buc, F., Fox, E.B., Garnett, R. (eds.)
  Advances in Neural Information Processing Systems 32: Annual Conference on
  Neural Information Processing Systems 2019, NeurIPS 2019, December 8-14,
  2019, Vancouver, BC, Canada. pp. 8024--8035 (2019),
  \url{https://proceedings.neurips.cc/paper/2019/hash/bdbca288fee7f92f2bfa9f7012727740-Abstract.html}

\bibitem{pearl_probabilistic_1988}
Pearl, J.: Probabilistic {Reasoning} in {Intelligent} {Systems}: {Networks} of
  {Plausible} {Inference}. Morgan Kaufmann Publishers Inc., San Francisco, CA,
  USA (1988)

\bibitem{scarselli_graph_2009}
Scarselli, F., Gori, M., Tsoi, A.C., Hagenbuchner, M., Monfardini, G.: The
  graph neural network model. {IEEE} Trans. Neural Networks  \textbf{20}(1),
  61--80 (2009). \doi{10.1109/TNN.2008.2005605},
  \url{https://doi.org/10.1109/TNN.2008.2005605}

\bibitem{selsam_learning_2019}
Selsam, D., Lamm, M., B{\"{u}}nz, B., Liang, P., de~Moura, L., Dill, D.L.:
  Learning a {SAT} solver from single-bit supervision. In: 7th International
  Conference on Learning Representations, {ICLR} 2019, New Orleans, LA, USA,
  May 6-9, 2019. OpenReview.net (2019),
  \url{https://openreview.net/forum?id=HJMC\_iA5tm}

\bibitem{SM19}
Soos, M., Meel, K.S.: {BIRD:} engineering an efficient {CNF-XOR} {SAT} solver
  and its applications to approximate model counting. In: The Thirty-Third
  {AAAI} Conference on Artificial Intelligence, {AAAI} 2019, The Thirty-First
  Innovative Applications of Artificial Intelligence Conference, {IAAI} 2019,
  The Ninth {AAAI} Symposium on Educational Advances in Artificial
  Intelligence, {EAAI} 2019, Honolulu, Hawaii, USA, January 27 - February 1,
  2019. pp. 1592--1599. {AAAI} Press (2019).
  \doi{10.1609/aaai.v33i01.33011592},
  \url{https://doi.org/10.1609/aaai.v33i01.33011592}

\bibitem{thurley_sharpsat_2006}
Thurley, M.: sharpsat - counting models with advanced component caching and
  implicit {BCP}. In: Biere, A., Gomes, C.P. (eds.) Theory and Applications of
  Satisfiability Testing - {SAT} 2006, 9th International Conference, Seattle,
  WA, USA, August 12-15, 2006, Proceedings. Lecture Notes in Computer Science,
  vol.~4121, pp. 424--429. Springer (2006). \doi{10.1007/11814948\_38},
  \url{https://doi.org/10.1007/11814948\_38}

\bibitem{valiant_complexity_1979}
Valiant, L.G.: The {Complexity} of {Enumeration} and {Reliability} {Problems}.
  SIAM J. Comput.  \textbf{8}(3),  410--421 (1979), publisher: Society for
  Industrial and Applied Mathematics

\bibitem{vaswani_attention_2017}
Vaswani, A., Shazeer, N., Parmar, N., Uszkoreit, J., Jones, L., Gomez, A.N.,
  Kaiser, L., Polosukhin, I.: Attention is all you need. In: Guyon, I., von
  Luxburg, U., Bengio, S., Wallach, H.M., Fergus, R., Vishwanathan, S.V.N.,
  Garnett, R. (eds.) Advances in Neural Information Processing Systems 30:
  Annual Conference on Neural Information Processing Systems 2017, December
  4-9, 2017, Long Beach, CA, {USA}. pp. 5998--6008 (2017),
  \url{https://proceedings.neurips.cc/paper/2017/hash/3f5ee243547dee91fbd053c1c4a845aa-Abstract.html}

\bibitem{velickovic_graph_2018}
Velickovic, P., Cucurull, G., Casanova, A., Romero, A., Li{\`{o}}, P., Bengio,
  Y.: Graph attention networks. In: 6th International Conference on Learning
  Representations, {ICLR} 2018, Vancouver, BC, Canada, April 30 - May 3, 2018,
  Conference Track Proceedings. OpenReview.net (2018),
  \url{https://openreview.net/forum?id=rJXMpikCZ}

\bibitem{sampleapprox}
Wang, J., Yin, M., Wu, J.: Two approximate algorithms for model counting.
  Theor. Comput. Sci.  \textbf{657},  28--37 (2017).
  \doi{10.1016/j.tcs.2016.04.047},
  \url{https://doi.org/10.1016/j.tcs.2016.04.047}

\bibitem{approxcount}
Wei, W., Selman, B.: A new approach to model counting. In: Bacchus, F., Walsh,
  T. (eds.) Theory and Applications of Satisfiability Testing, 8th
  International Conference, {SAT} 2005, St. Andrews, UK, June 19-23, 2005,
  Proceedings. Lecture Notes in Computer Science, vol.~3569, pp. 324--339.
  Springer (2005). \doi{10.1007/11499107\_24},
  \url{https://doi.org/10.1007/11499107\_24}

\bibitem{yolcu_learning_2019}
Yolcu, E., P{\'{o}}czos, B.: Learning local search heuristics for boolean
  satisfiability. In: Wallach, H.M., Larochelle, H., Beygelzimer, A.,
  d'Alch{\'{e}}{-}Buc, F., Fox, E.B., Garnett, R. (eds.) Advances in Neural
  Information Processing Systems 32: Annual Conference on Neural Information
  Processing Systems 2019, NeurIPS 2019, December 8-14, 2019, Vancouver, BC,
  Canada. pp. 7990--8001 (2019),
  \url{https://proceedings.neurips.cc/paper/2019/hash/12e59a33dea1bf0630f46edfe13d6ea2-Abstract.html}

\end{thebibliography}

\appendix 

\section{Extended Background}\label{app:back}

\subsection{Factor Graphs and Belief Propagation}\label{subapp:bp}

Belief Propagation (BP) \cite{pearl_probabilistic_1988} is an approximate inference algorithm that leverages the factor graph arising from its input probability distribution in its computational scheme. Given a probability distribution such as the one of Equation \ref{eq:prob}, the corresponding factor graph is a bipartite graph in which nodes are partitioned into factor nodes $F=\{f_1,\ldots, f_m\}$ (representing the factors of the probability distribution) and variable nodes $V=\{x_1, \ldots , x_n\}$ (representing the variables of the probability distribution). There is an undirected edge between a variable node $x_i\in V$ and a factor node $f_j\in F$ if and only if $f_j$ depends on $x_i$. Using this formalism, the neighbors $\mathcal{N}(f_j)$ of a factor node $f_j\in F$ are exactly all the variables appearing in $f_j$, and the neighbors $\mathcal{N}(x_i)$ of a variable node $x_i\in V$ are exactly all the factors that depend on that variable. 

Each node passes messages to its neighbors, so that variables send messages to factor nodes, and viceversa factors send messages to variable nodes, using the the iterative rules detailed in Equation \ref{eq:bp}.

It is possible to prove that BP is exact on tree factor graphs (i.e. it returns an exact solution in time linear in the number of variables $n$), however it can be very effective also on loopy graphs. When applied to graphs with loops, it is called Loopy Belief Propagation (LBP). 

After $T$ iterations of message passing (Equation \ref{eq:bp}), variables' beliefs (i.e. approximate marginals) $\{b_i(x_i)\}_{i=1}^n$ are computed as:

\begin{equation}
\begin{split}
b_i(x_i) &= \frac{1}{Z_i} \prod_{f_j\in \mathcal{N}(x_i)}m^{(T)}_{j\rightarrow i}(x_i) \\ Z_i &= \sum_{x_i} \prod_{f_j\in \mathcal{N}(x_i)} m_{j\rightarrow i}^{(T)}(x_i) 
\end{split}
\label{eq:beliefs_var}
\end{equation}
while factors' beliefs $\{b_j(x_{\mathcal{N}(f_j)})\}_{j=1}^m$ are computed as:

\begin{equation}
\begin{split}
   b_j(x_{\mathcal{N}(f_j)}) &= \frac{f_j(x_{\mathcal{N}(f_j)})}{Z_j} \prod_{x_i\in \mathcal{N}(f_j)} m^{(T)}_{i\rightarrow j}(x_i) \\
   Z_j &= \sum_{x_{\mathcal{N}(f_j)}} f_j(x_{\mathcal{N}(f_j)}) \prod_{x_i\in \mathcal{N}(f_j)} m_{i\rightarrow j}^{(T)}  
\end{split}
\label{eq:beliefs_fac}
\end{equation}
Using results of Equations \ref{eq:beliefs_var} and \ref{eq:beliefs_fac}, it is possible to compute the Bethe Free Energy $F$ of Equation \ref{eq:partition}.

In order to avoid underflow errors when implementing BP, message updates of Equation \ref{eq:bp} are usually performed in log-space, so that they read as follows:

\begin{equation}
\begin{split}
\hat{m}^{(k+1)}_{i\rightarrow j} (x_i) &= \sum_{c\in \mathcal{N}(x_i)\backslash j} \hat{m}^{(k)}_{c\rightarrow i}(x_i) \\
\hat{m}^{(k+1)}_{j\rightarrow i} (x_i) &= \text{LSE}_{x_1,\ldots, x_k\in \mathcal{N}(f_j)\backslash x_i} \bigg( f_j(x_i, x_1,\ldots, x_k) + \\ & +\sum_{x_v\in \mathcal{N}(f_j)\backslash x_i)} \hat{m}_{v\rightarrow i}^{(k)}(x_v) \bigg)
\end{split}
\label{eq:log_bp}
\end{equation}
where LSE stands for log-sum-exp function, defined as:
\begin{equation}
    \text{LSE}_{x_j\backslash x_i}(f_j(x_j)) = \ln \big(\sum_{x_j\backslash x_i} \exp (f_j(x_j))\big)
\end{equation}

\subsection{Graph Neural Networks}\label{subapp:gnn}

Graph Neural Networks (GNNs) \cite{scarselli_graph_2009} are a class of deep learning models that natively handle graph-structured data, with the goal of learning, for each vertex $v$ in the input graph $G=(V,E)$, a state embedding $\textbf{h}_v\in \mathbb{R}^{d}$, encoding information coming from neighboring nodes. The Message Passing Neural Network (MPNN) model \cite{gilmer_neural_2017} provides a general framework for GNNs, abstracting commonalities between many existing approaches in the literature. As the name suggests, the defining characteristic of MPNNs is that they use a form of neural message passing in which real-valued vector messages are exchanged between nodes (in particular between $1$-hop neighboring vertices), for a fixed number of iterations.

In detail, during each message passing iteration $t$ in a MPNN, the hidden representation $\textbf{h}_v^{(t)}$ of each node $v$ is updated according to the information $\textbf{m}_v^{(t)}$ aggregated from its neighborhood $\mathcal{N}(v)$ as:

\begin{equation}
\begin{split}
\textbf{m}_v^{(t+1)} & = \sum_{w\in \mathcal{N}(v)} M_t (\textbf{h}_v^{(t)}, \textbf{h}_w^{(t)}) \\
\textbf{h}_v^{(t+1)} & = U_t(\textbf{h}_v^{(t)}, \textbf{m}_v^{(t+1)})
\end{split}
\label{eq:mpnn_update}
\end{equation}
where $M_t$ is called message function and $U_t$ is called message update function: both are arbitrary differentiable functions (typically implemented as neural networks).

After running $T$ iterations of this message passing procedure, a readout phase is executed to compute the final embedding $\textbf{y}_v$ of each node $v$ as:

\begin{equation}
\textbf{y}_v = R(\{\textbf{h}_v^{(T)} | v\in G\})
\label{eq:mpnn_readout}
\end{equation}
where $R$ (the readout function) needs again to be a differentiable function, invariant to permutations of the node states.

Hence, the intuition behind MPNNs is that at each iteration every node aggregates information from its immediate graph neighbors, so as iterations progress each node embedding contains information from further reaches of the graph.  

Our model can be regarded as a MPNN taking as input the factor graph representing the input CNF SAT formula, whose message passing phase is the one detailed in Equations \ref{eq:var2fact} and \ref{eq:fact2var}, and the readout step is that of Equation \ref{eq:bpnn_readout}. It is worth noting that the distinction between factor and variable nodes makes the input graph heterogeneous. This is reflected into the MPNN architecture by using different update functions for factor and variable node embeddings, hence splitting each message passing iteration into two different phases: first every factor receives messages from its neighboring variables, then every variable receives messages from its neighboring factors. 
 
\section{Comparison with BPNN}\label{app:comp_bpnn}
In order to experimentally show the performance improvement given by augmenting the BPNN architecture (detailed in Section \ref{subsec:bpnn}) with a GAT-style attention mechanism, we also implemented and tested BPNN. The two MLPs of Equation \ref{eq:bpnn_update} are implemented as $3$-layer feedforward network, with ReLU activation between hidden layers; $\text{MLP}_3$ of Equation 9 and the learned operator $\Delta$ that transforms factor-to-variable messages are the same as BPGAT, detailed in Section \ref{subsec:bpgat_train}.

\begin{figure*}
    \centering
    \includegraphics[width=\textwidth, keepaspectratio]{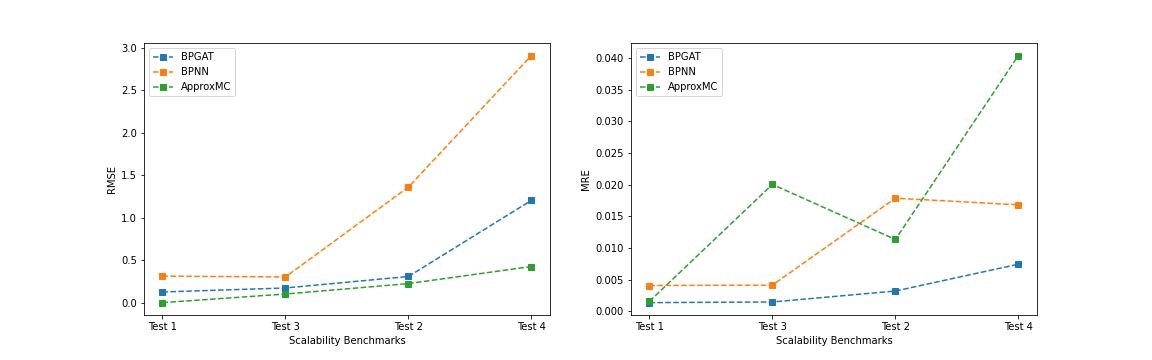}
    \caption{Summary of the results of the experiments done on BPNN and BPGAT to test their scalability. Notation is the same of Table \ref{tab:bpagt_bpnn}.}
    \label{fig:sum_scala}
\end{figure*}

Table \ref{tab:bpagt_bpnn} shows the results of testing both BPNN and BPGAT on the test sets described in Section 4.2, built to evaluate the scalability of the models. For all the benchmarks, BPGAT outperforms BPNN both in terms of RMSE and of MRE. A summary of these experiments is also reported on Figure \ref{fig:sum_scala}.

\begin{table}
\caption{RMSE/MRE comparison between BPGAT and BPNN on datasets of Boolean random formulae.}
\centering
\begin{tabular}{lrr}
\toprule
Dataset  & BPGAT  & BPNN \\
\midrule
Test 1    & 0.1276/0.001366 & 0.3140/0.004072     \\
Test 2    & 0.3100/0.003201  & 1.3623/0.01786      \\
Test 3    & 0.1748/0.001471  & 0.3046/0.004131      \\
Test 4    & 1.2061/0.007433  & 2.9112/0.01681 \\
\bottomrule
\end{tabular}
\label{tab:bpagt_bpnn}
\end{table}

We also performed a comparison of the generalization capabilities of BPNN and BPGAT under different training regimes, namely:
\begin{itemize}
    \item We fine-tuned the two models following the protocol described in Section \ref{subsec:bpgat_train}. Results are shown in Table \ref{tab:gener_ft}, where FT\_BPGAT and FT\_BPNN denote the fine-tuned BPGAT and BPNN architectures, respectively;
    \item We trained, for every distribution, the two models from scratch for $500$ epochs, with $250$ labeled examples. Results are shown in Table \ref{tab:gener_ts}, where TS\_BPGAT and TS\_BPNN denote the specifically-trained BPGAT and BPNN, respectively;
    \item We tested both BPGAT and BPNN trained on random Boolean formulae (with the training protocol and the data generating procedures described in Section \ref{subsec:bpgat_train}). Results are shown in Table \ref{tab:gener_rand}.
\end{itemize}

\begin{figure*}
    \centering
    \includegraphics[width=\textwidth, keepaspectratio]{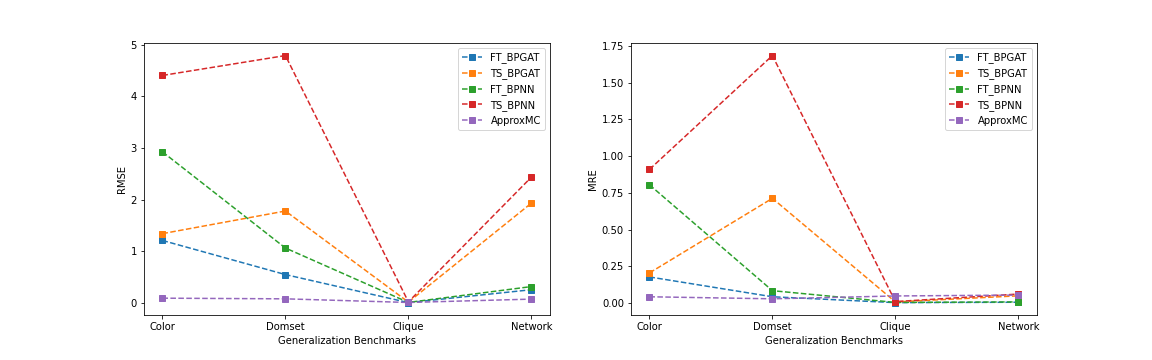}
    \caption{Summary of the results of the experiments done on BPNN and BPGAT to test their generalization capabilities. Notation is the same of Tables \ref{tab:gener_ts} and \ref{tab:gener_rand}.}
    \label{fig:sum_gene}
\end{figure*}

It is worth noting that, in all scenarios, BPGAT outperforms BPNN both in terms of RMSE and MRE. Moreover it is possible to observe that BPNN undergoes the same trend as BPGAT (Section \ref{subsec:results}): fine-tuning the architecture (pre-trained on random formulae) gives a better performance than training the model from scratch, using distribution-specific datasets.  A summary of these experiments is also reported on Figure \ref{fig:sum_gene}.

\begin{table}
\caption{RMSE/MRE comparison of BPGAT and BPNN fine-tuned for the specific distribution.}
\label{tab:gener_ft}
\centering
\begin{tabular}{lrr}
\toprule
Dataset  & FT\_BPGAT & FT\_BPNN \\
\midrule
Network    & 0.2580/0.005271  & 0.3187/0.007469 \\
Domset    & 0.5508/0.04252  & 1.0646/0.08392 \\
Color    & 1.2110/0.1774  & 2.9254/0.8046  \\
Clique    & 0.007834/0.002475   & 0.01201/0.004069  \\
\bottomrule
\end{tabular}
\end{table}

\begin{table}
\caption{RMSE/MRE comparison of BPGAT and BPNN trained with formulae sampled from the specific distribution.}
\label{tab:gener_ts}
\centering
\begin{tabular}{lrr}
\toprule
Dataset  & TS\_BPGAT & TS\_BPNN \\
\midrule
Network    & 1.9334/0.04887  & 2.4358/0.0610 \\
Domset    & 1.7808/0.7125  & 4.7885/1.6839 \\
Color    & 1.3430/0.2046  & 4.4036/0.9088  \\
Clique    & 0.01773/0.007333   & 0.02265/0.009737 \\
\bottomrule
\end{tabular}
\end{table}

\begin{table}
\caption{RMSE/MRE comparison of BPGAT and BPNN trained on random formulae.}
\label{tab:gener_rand}
\centering
\begin{tabular}{lrr}
\toprule
Dataset  & BPGAT & BPNN \\
\midrule
Network    & 14.2839/0.3608  & 33.0484/0.8328 \\
Domset    & 11.9190/9.5070  & 38.7884/13.3770 \\
Color    & 26.1898/5.9593  & 36.1179/7.8826   \\
Clique    & 1.9625/0.8983   & 5.2792/2.4177  \\
\bottomrule
\end{tabular}
\end{table}

\section{Configuration and Ablation Studies}\label{app:ablations}

\subsection{BPGAT Parameter Tuning}\label{subapp:par_tuning}

Before setting BPGAT architecture's parameters to the ones detailed in Section \ref{subsec:bpgat_train}, we performed several tuning experiments. 

\paragraph{Learning the Damping Parameter} In order to assess whether it is better to use a fixed scalar parameter $\alpha$ or a learned operator $\Delta$ as damping parameter of Equation \ref{eq:damping}, we tried all possible configurations (each configuration is trained as specified in Section \ref{subsec:bpgat_train}). 

In Table \ref{tab:damp_conf}, BPGAT refers to the architecture with damping parameter computed by an MLP $\Delta$ for factor-to-variable messages, and fixed to $0.5$ for variable-to-factor messages; BPGAT\_VF refers to the architecture with damping parameter computed by an MLP $\Delta$ for variable-to-factor messages, and fixed to $0.5$ for factor-to-variable messages; BPGAT\_ALL refers to the model trained with an MLP $\Delta$ for transforming both kind of messages; BPGAT\_NONE refers to the model trained with both damping parameter fixed to $0.5$. Results justify our choice of using a learned operator $\Delta$ only for updating factor-to-variable messages.

\begin{table*}
\caption{RMSE/MRE comparison of BPGAT trained with different configurations of the damping parameter.}
\label{tab:damp_conf}
\centering
\begin{tabular}{lrrrr}
\toprule
Dataset  & BPGAT  & BPGAT\_VF & BPGAT\_ALL & BPGAT\_NONE \\
\midrule
Test 1    & 0.1276/0.001366 & 0.2351/0.002762 & 0.3157/0.003930 & 0.4158/0.006443     \\
Test 2    & 0.3100/0.003201  & 0.8525/0.01001 & 0.9493/0.01601 & 0.9238/0.01555      \\
Test 3    & 0.1748/0.001471  & 0.2109/0.002998 & 0.2589/0.003041 & 0.3013/0.003344    \\
\bottomrule
\end{tabular}
\end{table*}

\paragraph{Number of Message Passing Iterations} Another parameter that we explored is the number $T$ of message passing rounds, i.e. the number of times Equations \ref{eq:var2fact} and \ref{eq:fact2var} are iterated before the readout phase (Equation \ref{eq:bpnn_readout}) is executed. Results are shown in Table \ref{tab:n_iters}: the model achieves the highest performance when $T=5$. 

\begin{table*}
\caption{RMSE/MRE comparison for different number of message passing iterations $T$. BPGAT, BPGAT\_10 and BPGAT\_15 refer to the model trained with $T=5, 10, 15$, respectively.}
\label{tab:n_iters}
\centering
\begin{tabular}{lrrrr}
\toprule
Dataset  & BPGAT  & BPGAT\_10 & BPGAT\_15 \\
\midrule
Test 1    & 0.1276/0.001366 & 0.1684/0.003415 & 0.1775/0.003694     \\
Test 2    & 0.3100/0.003201  & 0.3751/0.005622 & 0.5328/0.006294      \\
Test 3    & 0.1748/0.001471  & 0.2555/0.003165 & 0.2784/0.005979    \\
\bottomrule
\end{tabular}
\end{table*}

\subsection{Ablation Studies on GAT Layers}\label{subapp:ablat}

To assess the relevance of using a GAT-style attention mechanism in transforming both factor-to-variable and variable-to-factor messages, we performed the following ablation studies:
\begin{itemize}
    \item Hybrid BP-BPGAT, results are shown in Table \ref{tab:att_abl}: we tested the model transforming factor-to-variable messages using GAT (Equation \ref{eq:fact2var}) and variable-to-factor messages using BP (Equation \ref{eq:bp}), this is denoted as FVGAT-VFNONE in Table \ref{tab:att_abl}; conversely, we tested the model transforming variable-to-factor messages using BP (Equation \ref{eq:bp}) and factor-to-variable messages using GAT (Equation \ref{eq:var2fact}), this is denoted as FVNONE-VFGAT in Table \ref{tab:att_abl};
    \item Hybrid BPNN-BPGAT, results are shown in Table \ref{tab:abl_att}: we tested the model transforming factor-to-variable messages using GAT (Equation \ref{eq:fact2var}) and variable-to-factor messages using BPNN's corresponding update (Equation \ref{eq:bpnn_update}), this is denoted as FVGAT-VFMLP in Table \ref{tab:abl_att}; conversely, we tested the model transforming variable-to-factor messages using BPNN's update (Equation \ref{eq:bpnn_update}) and factor-to-variable messages using GAT (Equation \ref{eq:var2fact}), this is denoted as FVMLP-VFGAT in Table \ref{tab:abl_att};
\end{itemize}

Results highlight the fact that transforming both factor-to-variable and variable-to-factor messages using a GAT attention mechanism is beneficial for the overall performance of the model. 

\begin{table*}
\caption{RMSE/MRE results of the ablation studies on the attention mechanism.}
\label{tab:att_abl}
\centering
\begin{tabular}{lrrr}
\toprule
Dataset  & BPGAT  & FVGAT-VFNONE  & FVNONE-VFGAT \\
\midrule
Test 1    & 0.1276/0.001366 & 0.1568/0.0015721 & 0.1786/0.001740     \\
Test 2    & 0.3100/0.003201  & 0.4292/0.004640 & 0.5870/0.006392      \\
Test 3    & 0.1748/0.001471  & 0.2009/0.001964 & 0.2699/0.002197      \\
\bottomrule
\end{tabular}
\end{table*}

\begin{table*}
\caption{RMSE/MRE results of the ablation studies between message transformation performed by BPGAT and BPNN.}
\label{tab:abl_att}
\centering
\begin{tabular}{lrrr}
\toprule
Dataset  & BPGAT  & FVGAT-VFMLP  & FVMLP-VFGAT \\
\midrule
Test 1    & 0.1276/0.001366 & 0.3613/0.004713 & 0.4056/0.005256     \\
Test 2    & 0.3100/0.003201  & 1.5787/0.02002 & 1.7065/0.02211      \\
Test 3    & 0.1748/0.001471  & 0.2751/0.004689 & 0.2902/0.004872      \\
\bottomrule
\end{tabular}
\end{table*}

\end{document}


\maketitle

\appendix 

\section{Extended Background}

\section{Comparison with BPNN}

In order to experimentally show the performance improvement given by augmenting the BPNN architecture (detailed in Section 3.1) with a GAT-style attention mechanism, we implemented and tested it. The two MLPs of Equation 8 are implemented as $3$-layer feedforward network, with ReLU activation between hidden layers; $\text{MLP}_3$ of Equation 9 and the learned operator $\Delta$ that transforms factor-to-variable messages are the same as BPGAT, detailed in Section 4.1.

Table \ref{tab:bpagt_bpnn} shows the results of testing both BPNN and BPGAT on some of the test sets described in Section 4.2, built to evaluate the scalability of the models. For all the benchmarks, BPGAT outperforms BPNN both in terms of RMSE and of MRE.

\begin{table}
\centering
\begin{tabular}{lrr}
\toprule
Dataset  & BPGAT  & BPNN \\
\midrule
Test Set 1    & 0.1276/0.001366 & 0.3140/0.004072     \\
Test Set 2    & 0.3100/0.003201  & 1.3623/0.01786      \\
Test Set 3    & 0.1748/0.001471  & 0.3046/0.004131      \\
\bottomrule
\end{tabular}
\caption{RMSE/MRE comparison between BPGAT and BPNN}
\label{tab:bpagt_bpnn}
\end{table}

Table \ref{tab:gener_rand} shows the results of the experiments done on the datasets used to test the generalization capabilities of the model (described in Section 4.2), both for BPGAT and BPNN. 

\begin{table}
\centering
\begin{tabular}{lrr}
\toprule
Dataset  & BPGAT & BPNN \\
\midrule
Network    & 14.2839/0.3608  & 33.0484/0.8328 \\
Domset    & 11.9190/9.5070  & 38.7884/13.3770 \\
Color    & 26.1898/5.9593  & 36.1179/7.8826   \\
Clique    & 1.9625/0.8983   & 5.2792/2.4177  \\
\bottomrule
\end{tabular}
\caption{RMSE/MRE comparison of BPGAT and BPNN trained on random formulae.}
\label{tab:gener_rand}
\end{table}

\begin{table}
\centering
\begin{tabular}{lrr}
\toprule
Dataset  & FT\_BPGAT & FT\_BPNN \\
\midrule
Network    & 0.2580/0.005271  & 0.3187/0.007469 \\
Domset    & 0.5508/0.04252  & 1.0646/3.5070 \\
Color    & 1.2110/0.1774  & 2.9254/0.8046  \\
Clique    & 0.007834/0.002475   & 0.022651/0.009737  \\
\bottomrule
\end{tabular}
\caption{RMSE/MRE comparison of BPGAT and BPNN fine-tuned for the specific distribution.}
\label{tab:gener_ft}
\end{table}

\begin{table}
\centering
\begin{tabular}{lrr}
\toprule
Dataset  & TS\_BPGAT & TS\_BPNN \\
\midrule
Network    & 1.9334/0.04887  & 2.4358/0.0610 \\
Domset    & 1.7808/0.7125  & 4.7885/1.6839 \\
Color    & 1.3430/0.2046  & 4.4036/0.9088  \\
Clique    & 0.01773/0.007333   & 0.02265/0.009737 \\
\bottomrule
\end{tabular}
\caption{RMSE/MRE comparison of BPGAT and BPNN trained with formulae sampled from the specific distribution.}
\label{tab:gener_ts}
\end{table}

\section{Ablation Studies}

\begin{table*}
\centering
\begin{tabular}{lrrrr}
\toprule
Dataset  & BPGAT  & BPGAT\_VF & BPGAT\_ALL & BPGAT\_NONE \\
\midrule
Test Set 1    & 0.1276/0.001366 & 0.2351/0.002762 & 0.3157/0.003930 & 0.3158/0.008443     \\
Test Set 2    & 0.3100/0.003201  & 0.8525/0.01001 & 0.9493/0.01601 & 0.9238/0.01555      \\
Test Set 3    & 0.1748/0.001471  & 0.2109/0.002998 & 0.2589/0.003041 & 0.3013/0.003344    \\
\bottomrule
\end{tabular}
\caption{RMSE/MRE comparison for different configurations of the damping parameter.}
\label{tab:damp_conf}
\end{table*}

\begin{table*}
\centering
\begin{tabular}{lrrr}
\toprule
Dataset  & BPGAT  & FVGAT-VFNONE  & FVNONE-VFGAT \\
\midrule
Test Set 1    & 0.1276/0.001366 & 0.1568/0.0015721 & 0.1786/0.001740     \\
Test Set 2    & 0.3100/0.003201  & 0.4292/0.004640 & 0.5870/0.006392      \\
Test Set 3    & 0.1748/0.001471  & 0.2009/0.001964 & 0.2699/0.002197      \\
\bottomrule
\end{tabular}
\caption{RMSE/MRE results of the ablation studies about the attention mechanism.}
\label{tab:att_abl}
\end{table*}

\begin{table*}[t]
\centering
\begin{tabular}{lrrr}
\toprule
Dataset  & BPGAT  & FVGAT-VFMLP  & FVMLP-VFGAT \\
\midrule
Test Set 1    & 0.1276/0.001366 & 0.3613/0.004713 & 0.4056/0.005256     \\
Test Set 2    & 0.3100/0.003201  & 1.5787/0.02002 & 1.7065/0.02211      \\
Test Set 3    & 0.1748/0.001471  & 0.2751/0.004689 & 0.2902/0.004872      \\
\bottomrule
\end{tabular}
\caption{RMSE/MRE results of the ablation studies between message transformation performed by BPGAT and BPNN.}
\label{tab:abl_att}
\end{table*}